\documentclass{article}

\usepackage{microtype}
\usepackage{graphicx}
\usepackage{subcaption}
\usepackage{booktabs} % for professional tables
\usepackage{graphicx} % Required for inserting images
\usepackage{multirow} % For merging rows
\usepackage{xcolor}   % For colored text
\usepackage{booktabs} % Recommended for professional looking tables
\usepackage{colortbl}
\usepackage{tabularx}
\usepackage{subcaption} 
\usepackage{makecell}
\usepackage{xspace}

\usepackage{hyperref}

\usepackage[preprint]{icml2026}

\renewcommand{\headrulewidth}{0.4pt}

\usepackage{amsmath}
\usepackage{amssymb}
\usepackage{mathtools}
\usepackage{amsthm}
\usepackage{lipsum}
\usepackage{xcolor}
\usepackage{enumitem}
\usepackage{bbm}
\usepackage[capitalize,noabbrev]{cleveref}

\theoremstyle{plain}

\theoremstyle{definition}

\theoremstyle{remark}

\newcommand{\methodname}{\textsc{PBio-Agent}}
\newcommand{\benchmarkname}{\textsc{LincsQA}}

\newcommand{\ie}{\emph{i.e.},\@\xspace}
\newcommand{\eg}{\emph{e.g.},\@\xspace}

\usepackage[textsize=tiny]{todonotes}

\definecolor{denim}{rgb}{0.08, 0.38, 0.74}
\definecolor{darkspringgreen}{rgb}{0.09, 0.45, 0.27}
\definecolor{darkbrown}{rgb}{0.4, 0.26, 0.13}

\hypersetup{
    colorlinks=true,
    linkcolor=denim,
    citecolor=denim,
    urlcolor=denim
}
\icmltitlerunning{Submission and Formatting Instructions for ICML 2026}

\begin{document}

\twocolumn[
  \icmltitle{Progressive Multi-Agent Reasoning for Biological Perturbation Prediction}
% Hyomin Kim, Sang-Yeon Hwang, Jaechang Lim, Yinhua Piao, Yunhak Oh, Woo Youn Kim, Chanyoung Park, Sungsoo Ahn, Junhyeok Jeon
  \icmlsetsymbol{equal}{*}

  \begin{icmlauthorlist}
    \icmlauthor{Hyomin Kim}{KAIST,HITS}
    \icmlauthor{Sang-Yeon Hwang}{HITS}
    \icmlauthor{Jaechang Lim}{HITS}
    \icmlauthor{Yinhua Piao}{KAIST}
    \icmlauthor{Yunhak Oh}{KAIST,HITS}
    \icmlauthor{Woo Youn Kim}{HITS}
    \icmlauthor{Chanyoung Park}{KAIST}
    \icmlauthor{Sungsoo Ahn}{KAIST}
    \icmlauthor{Junhyeok Jeon}{equal,HITS}
  \end{icmlauthorlist}

  \icmlaffiliation{KAIST}{Kim Jaechul Graduate School of AI, Seoul, Korea}
  \icmlaffiliation{HITS}{HITS, Seoul, Korea}
  \icmlcorrespondingauthor{Junhyeok Jeon}{jhjeon@hits.ai}
  \icmlkeywords{Machine Learning, ICML}

  \vskip 0.3in
]
\printAffiliationsAndNotice{Work conducted during HITS internship by H. Kim and Y. Oh.} 

\begin{abstract}
Predicting gene regulation responses to biological perturbations requires reasoning about underlying biological causalities. While large language models (LLMs) show promise for such tasks, they are often overwhelmed by the entangled nature of high-dimensional perturbation results. Moreover, recent works have primarily focused on genetic perturbations in single-cell experiments, leaving bulk-cell chemical perturbations, which is central to drug discovery, largely unexplored.
Motivated by this, we present \benchmarkname, a novel benchmark for predicting target gene regulation under complex chemical perturbations in bulk-cell environments. We further propose \methodname, a multi-agent framework that integrates difficulty-aware task sequencing with iterative knowledge refinement. Our key insight is that genes affected by the same perturbation share causal structure, allowing confidently predicted genes to contextualize more challenging cases. The framework employs specialized agents enriched with biological knowledge graphs, while a synthesis agent integrates outputs and specialized judges ensure logical coherence.
\methodname\ outperforms existing baselines on both \benchmarkname\ and PerturbQA, enabling even smaller models to predict and explain complex biological processes without additional training.
\end{abstract}
%\vspace{-4mm}

\begin{figure*}[t]
    \centering
    \includegraphics[width=\textwidth]{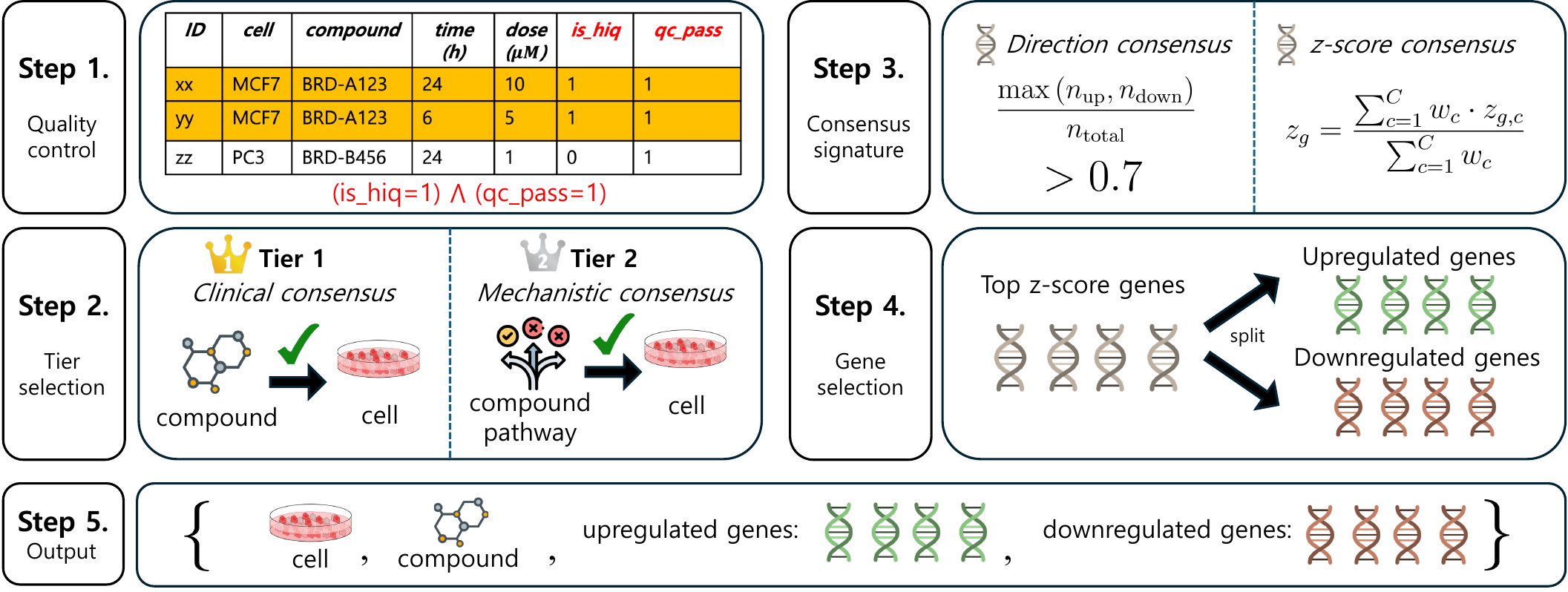}
    
    \caption{\textbf{Overview of the \benchmarkname\ benchmark construction.} (i) Quality control: Filtering LINCS L1000 Level 5 signatures for high-quality compound treatments. (ii) (b) Tier Selection: Hierarchical pairing of compounds to cell lines using a two-tier strategy. Tier 1 (clinical consensus) requires strict clinical indication alignment, where the compound's approved therapeutic use must match the cell line's disease origin. Tier 2 (mechanistic consensus) applies when no clinically matched cell line is available, selecting instead based on target biology and pathway activity relevant to the compound's mechanism, independent of clinical indication. (iii) Consensus signature: Extracting robust signals by enforcing directional consistency ($\geq 0.7$) and computing replicate-weighted consensus $z$-scores ($z_g$). (iv) Gene selection: Ranking and filtering genes by $z$-score magnitude and MoA-plausibility to form binary queries. (v) Output: Final benchmark comprising specific cell-compound contexts paired with high-confidence up- and down-regulated gene sets.}
    \label{fig:lincqa_overview}
\end{figure*}

\vspace{-9mm}

\section{Introduction}

Large language models have shown distinct benefits in biological reasoning and have led to their increasing use in perturbation biology \citep{martens2025langpert, wu2025perturbqa}. Recent efforts in the computational biology and machine learning communities have developed benchmarks \citep{youngblut2025scbasecount, zhang2025tahoe100m, wu2024perturbench, levine2024cell2sentence, wu2025perturbqa} and methods \citep{roohani2024gears, lu2025cellclip, adduri2025state, wenkel2025txpert} for predicting transcriptional responses to perturbations at the single-cell level.

While single-cell perturbation studies have garnered significant attention, bulk-cell chemical perturbations remain central to drug discovery. The LINCS L1000 database \citep{duan2016lincsl1000cds2} profiles over 20,000 compounds, far exceeding the genetic perturbation space of typical Perturb-seq \citep{dixit2016perturbseq} experiments, with well-annotated mechanisms of action that enable direct evaluation of mechanistic reasoning. Moreover, bulk signatures capture population-averaged responses that mirror tissue-level readouts in preclinical studies, where aggregate effects determine therapeutic outcomes. Despite this practical importance, no benchmark currently evaluates LLMs on bulk-cell chemical perturbation prediction.

To bridge this gap, we introduce \benchmarkname, a benchmark to evaluate how LLMs reason through biological responses to drugs in bulk cell. Inspired by PerturbQA \citep{wu2025perturbqa}, \benchmarkname\ evaluates gene regulation direction (up/down) prediction under a specific mechanism hypothesis. Crucially, we assess whether LLM predictions exhibit \emph{biological validity}: for a given compound and its known mechanism of action, the model's agreement with observed transcriptional changes should be significantly higher in cell lines where the drug is pharmacologically active (sensitive) than in cell lines where the target pathway is absent or bypassed (insensitive). This design directly tests whether LLMs understand that a drug's mechanism only manifests transcriptionally when the relevant biological context is present. Finally, \benchmarkname\ uses consensus signatures to filter out experimental noise, providing a stable ground truth for evaluating transcriptional changes.

In addition to the benchmark, we introduce \methodname, a multi-agent framework that decomposes biological reasoning across specialized agents with access to structured knowledge. Predicting gene regulation from chemical perturbations requires compositional reasoning: retrieving molecular targets, tracing cell-type-specific signaling cascades, and resolving conflicting regulatory signals. Monolithic LLMs frequently hallucinate pathway connections or ignore cellular context when these subtasks are handled implicitly. Our framework addresses this by assigning each subtask to dedicated agents, and further exploits the shared causal structure among genes affected by the same perturbation through progressive reasoning, as illustrated in \Cref{fig:method_overview}.

For a concrete example, when a MEK inhibitor is applied to melanoma cells, the model first predicts with high confidence that canonical ERK target genes such as FOS and EGR1 are downregulated. This prediction then provides information for a more challenging case: DUSP6, a phosphatase that dephosphorylates ERK as part of a negative feedback loop, is itself transcriptionally induced by ERK and should therefore also decrease upon MEK inhibition. However, LLMs often incorrectly predict upregulation by conflating the functional role of DUSP6 in opposing ERK activity with its transcriptional regulation by ERK. By ordering predictions from easy to hard and propagating context, our framework captures such mechanistic dependencies without additional training.

Our multi-agent system integrates specialized reasoning agents that leverage biological knowledge graphs and databases \citep{perfetto2016signor, killian2021depmap, huttlin2015bioplex, giurgiu2019corum, martin2023ensembl, uniprot2018uniprot, ashburner2000geneontologygo, szklarczyk2023string, milacic2024reactome} to trace regulatory pathways and identify molecular mechanisms. Optionally, the framework can incorporate pre-trained neural networks as tools, such as graph attention networks \citep{velivckovic2017gat} or random forests \citep{rigatti2017randomforest} for pathway inference, enabling ensemble predictions.

In summary, this work makes the following contributions:
% \vspace{-4mm}
\begin{itemize}[leftmargin=*, nosep] % nosep 옵션 추가
    \item \benchmarkname, the first comprehensive benchmark for evaluating LLM-based bulk-cell chemical perturbation prediction, featuring gene-level accuracy assessment and cell-line-context evaluation that tests biological validity across pharmacologically sensitive and insensitive conditions.
    \item \methodname, a training-free multi-agent system that significantly enhances prediction accuracy by leveraging the interconnected nature of perturbation-induced transcriptional changes.
\end{itemize}
\section{\benchmarkname\ benchmark}
\label{sec:method_lincqa_benchmark}
\benchmarkname\ (LINCS-based Question-Answering benchmark) evaluates LLM's ability to reason about transcriptional responses to compound perturbations. In this section, we describe construction process of \benchmarkname\ at  \Cref{sec:method_dataset_curation} and visualize at \Cref{fig:lincqa_overview}. Then we describe two tasks of \benchmarkname{} at \Cref{sec:method_tasks}. We further elaborate data statistics and metrics at \Cref{appdx:dataset} and \Cref{appdx:metrics} respectively.

\subsection{Dataset curation.} \label{sec:method_dataset_curation}

\paragraph{Dataset preparation.} 
Gene expression signatures were derived from the LINCS 2020 dataset level 5\footnote{\url{https://clue.io/releases/data-dashboard}}. We utilized \texttt{compoundinfo\_beta.txt} for compound–MoA mappings, \texttt{siginfo\_beta.txt} for signature metadata, and \texttt{compound\_signatures.h5} for precomputed z-score profiles. As described in step 1 of \Cref{fig:lincqa_overview}, signatures were filtered to retain high-quality perturbation data meeting the following criteria: (i) perturbation type \texttt{trt\_cp} (compound treatment), (ii) \texttt{is\_hiq} flag set to true, indicating "high-quality" signatures and (iii) \texttt{qc\_pass} flag set to true, confirming that the sample passed all standard laboratory quality control metrics. Only compounds with annotated MoA information were included, yielding a pool of 2,559 compounds with defined mechanisms.

\paragraph{Cell line selection.}
Cell line selection follows two task-dependent strategies: gene-level prediction and cell-line-context evaluations. For gene-level prediction, we employed a two-tier strategy to pair compounds with biologically relevant cell lines. The first-tier prioritized clinical context, using Gemini 2.0 Flash \citep{team2023gemini} to match a compound's therapeutic indication with a cell line's disease background. For compounds without direct disease matches, a second tier focused on mechanistic relevance was applied. The model was instructed to abstain if no appropriate match existed. After a secondary LLM validation of the clinical rationale, 193 high-confidence compound--cell line pairs were retained for the final benchmark.

For the cell-line-context evaluation, we manually curated cell lines based on molecular features and drug sensitivity. Sensitive cell lines were selected based on known responsive features and sub-micromolar IC$_{50}$ values (\eg, A375 melanoma cells carries the BRAF V600E mutation for BRAF inhibitors). Conversely, insensitive cell lines were chosen based on the absence of the drug target or the presence of bypass resistance mechanisms. These annotations were grounded in published IC$_{50}$ data and drug response literature. We further provide details in \Cref{exp:lincsqa_benchmark}.

\paragraph{Consensus signature construction.}
Transcriptional responses to compound perturbation exhibit substantial biological variability across experimental conditions, including differences in dose, exposure time, and replicate measurements. To address this inherent heterogeneity, \benchmarkname\ aggregates expression data across all available conditions for each compound--cell line pair, thereby reducing sensitivity to condition-specific artifacts associated with single-condition measurements.

For each gene, we computed a directional consistency score:
\begin{equation}
\text{Consistency}_g = 
\frac{\max\left(n_{\text{up}}, n_{\text{down}}\right)}{n_{\text{total}}}
\end{equation}
where $n_{\text{up}}$, $n_{\text{down}}$, and $n_{\text{total}}$ represent the number of conditions showing upregulation, downregulation, and total conditions, respectively. Genes with consistency scores greater than or equal to 0.7 were retained, ensuring robust directional agreement despite condition-specific noise.

For the retained genes, a consensus $z$-score was computed as a replicate-weighted average:
\begin{equation}
z_g =
\frac{\sum_{c=1}^{C} w_c \cdot z_{g,c}}
{\sum_{c=1}^{C} w_c}
\end{equation}
where $w_c$ denotes the number of replicate signatures for condition $c$. This consensus approach captures reproducible transcriptional signals while filtering stochastic variation, providing a robust foundation for benchmark query construction.

\paragraph{Query gene selection.}
Genes used for benchmark queries were selected through a two-stage process designed to balance biological relevance with LLM context constraints. In the first stage, genes that passed the consistency threshold were ranked by the absolute magnitude of their consensus $z$-scores. In the second stage, the top-ranked genes were further filtered based on mechanistic plausibility with respect to the annotated MoA, with up to ten genes selected per direction (upregulated and downregulated).
Each selected gene yielded a single binary query. The resulting query set typically comprised 10--20 genes per test case, subject to a minimum requirement of 40 consistently regulated genes per signature.

\subsection{Tasks} \label{sec:method_tasks}
 \paragraph{Gene-level regulation task.} Inspired by \citet{wu2025perturbqa} gene-level task evaluates the model's ability to predict the transcriptional response of specific genes following chemical perturbation. For a given compound--cell line pair, the model is queried to determine whether a target gene is upregulated or downregulated. This level focuses on the direct mapping between a compound's biochemical influence and its downstream effect on the expression of individual genes, serving as a fundamental test of the model’s mechanistic reasoning.

\paragraph{MoA-level context task.} 
To assess deeper biological validity, we evaluated the prediction accuracy across cell lines with divergent pharmacological sensitivities. Here, pharmacological sensitivity reflects the extent to which a compound is expected to elicit a functional response in a given cellular context, as determined by effective target engagement, pathway dependence, and the absence of dominant compensatory or bypass mechanisms. 
In sensitive cell lines, perturbation of the intended target is more likely to propagate through downstream signaling networks and give rise to coherent transcriptional changes, whereas in insensitive contexts such coupling may be weakened by low target expression, pathway redundancy, or adaptive feedback and homeostatic mechanisms that restore pathway output despite target inhibition.
We hypothesize that a transcriptional-response--based MoA prediction framework with genuine mechanistic grounding should exhibit higher predictive agreement with observed transcriptional changes in sensitive cell lines compared to insensitive ones. 
By measuring this performance gap, we aim to quantify the extent to which the model captures the interaction between a drug's mode of action and its cellular environment, rather than relying on context-agnostic correlations.

\begin{figure*}[t]
    \centering
    \begin{subfigure}[b]{0.48\textwidth}
        \centering
        \includegraphics[width=\textwidth]{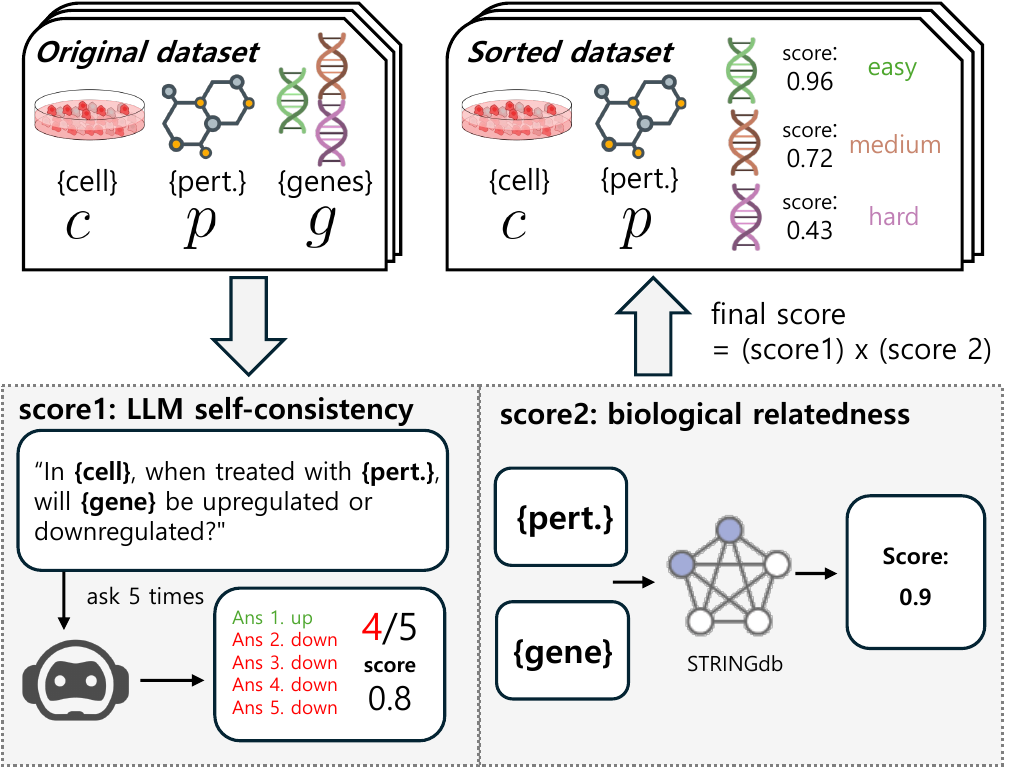} 
        \caption{Difficulty aware data sorting}
        \label{fig:difficulty_aware_data_ordering}
    \end{subfigure}
    \hfill
    \begin{subfigure}[b]{0.48\textwidth}
        \centering
        \includegraphics[width=\textwidth]{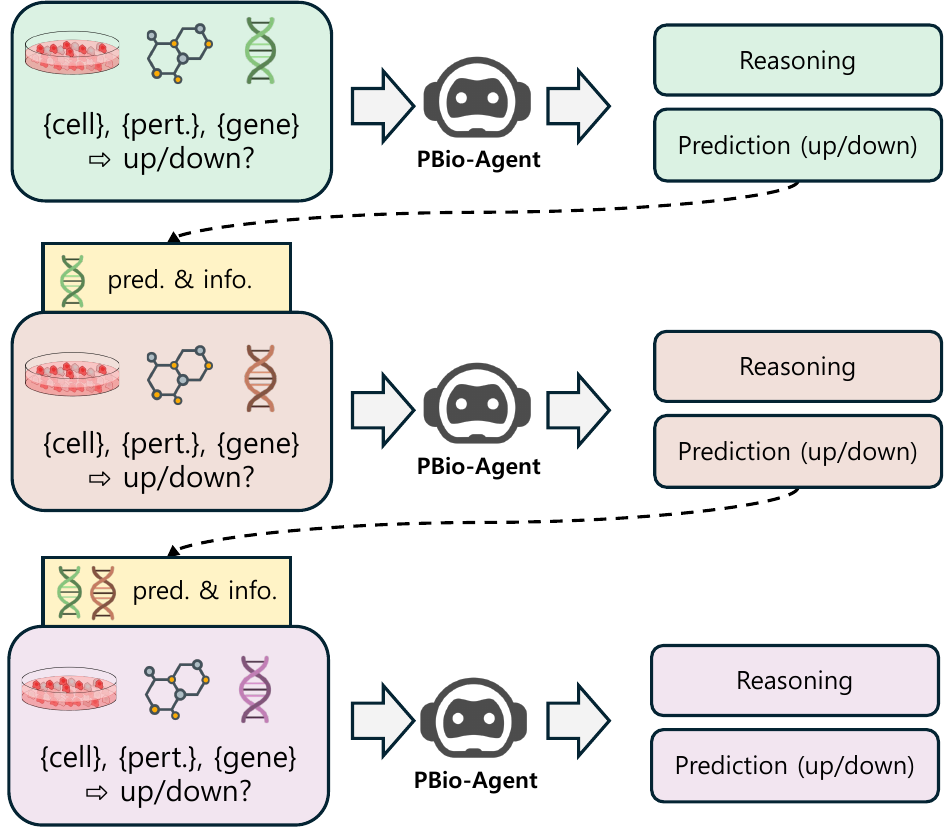} 
        \caption{Progressive reasoning}
        \label{fig:progressive_reasoning}
    \end{subfigure}
    
    \caption{\textbf{Overview of \methodname.} \textbf{(a) Difficulty aware data sorting:} We order data using a composite score derived from the product of two metrics. LLM self-consistency measures prediction stability over multiple trials. Biological relatedness of perturbation and gene is fetched from the STRING database.  \textbf{(b) Progressive reasoning:} \methodname\ processes genes from easy to hard to build iterative context. High confidence predictions and reasoning traces from earlier steps are propagated as supplementary information to guide the analysis of subsequent, more complex biological cases.}
    \label{fig:method_overview}
\end{figure*}
\section{\methodname} \label{sec:method_pbioagent}
\paragraph{Problem formulation.}
\methodname\ is a multi-agent framework that aims to predict transcriptional responses in a bulk-cell setting. Let $\mathcal{C}$, $\mathcal{P}$, and $\mathcal{G}$ denote the sets of cell lines, chemical perturbations (compounds), and target genes, respectively. We define a sample as a tuple $x = (c, p, g) \in \mathcal{C} \times \mathcal{P} \times \mathcal{G}$. The objective is to learn a mapping $f: \mathcal{X} \to \mathcal{Y}$ where $\mathcal{Y} = \{0, 1\}$ represents the binary direction of gene regulation (0 for downregulation, 1 for upregulation). We denote the dataset as $\mathcal{D} = \{(x_i, y_i)\}_{i=1}^N$. The model receives the perturbation metadata $x_i$ along with textual context and structured knowledge $\mathcal{K}$, aiming to maximize the likelihood $P(y_i | x_i, \mathcal{K})$ while respecting cell-line specificity.

\begin{table*}[t]
    \centering
    \small
    \renewcommand{\arraystretch}{1.2}
    \setlength{\tabcolsep}{0pt}
    \begin{tabular*}{\textwidth}{@{\extracolsep{\fill}} l *{8}{c} @{}}
    \toprule
    \textbf{Model} & \textbf{Bone marrow} & \textbf{Breast} & \textbf{Cervix} & \textbf{Colon} & \textbf{Lung} & \textbf{Periph. blood} & \textbf{Prostate} & \textbf{Skin} \\
    \midrule
    \rowcolor{gray!10}
    \multicolumn{9}{c}{\textit{Generalist LLM}} \\
    Llama3-8B                      &    0.46$\pm$.02     &    0.53$\pm$.02     &    0.49$\pm$.01     &    \underline{0.53$\pm$.07}     &    \underline{0.58$\pm$.02}     &    \underline{0.54$\pm$.05}     &    0.50$\pm$.04     &    0.53$\pm$.01     \\
    DeepSeek-R1-Distill-Llama-8B   &    \underline{0.49$\pm$.08}     &    0.49$\pm$.03     &    0.51$\pm$.04     &    0.46$\pm$.04     &    0.50$\pm$.04     &    \underline{0.54$\pm$.03}     &    0.52$\pm$.04     &    0.50$\pm$.02     \\
    Mistral-Small-3.2-24B          &    0.41$\pm$.05     &    0.52$\pm$.01     &    0.50$\pm$.02     &    0.47$\pm$.04     &    0.55$\pm$.02     &    0.51$\pm$.01     &    0.52$\pm$.01     &    0.53$\pm$.02     \\
    Qwen3-30B-A3B                  &    0.47$\pm$.03     &    \underline{0.62$\pm$.01}     &    \underline{0.53$\pm$.02}     &    0.43$\pm$.05     &    \underline{0.58$\pm$.01}     &    0.52$\pm$.02     &    \textbf{0.66$\pm$.07}     &    \underline{0.60$\pm$.02}     \\
    \midrule
    \rowcolor{gray!10}
    \multicolumn{9}{c}{\textit{Specialist LLM}} \\
    BioMistral-7B                  &    0.47$\pm$.02     &    0.51$\pm$.02     &    0.49$\pm$.01     &    0.51$\pm$.05     &    0.49$\pm$.01     &    0.49$\pm$.02     &    0.50$\pm$.03     &    0.51$\pm$.01     \\
    BioMedGPT-LM-7B                &    \underline{0.49$\pm$.01}     &    0.50$\pm$.00     &    0.50$\pm$.00     &    0.50$\pm$.00     &    0.50$\pm$.00     &    0.50$\pm$.00     &    0.50$\pm$.00     &    0.50$\pm$.01     \\
    TxGemma-27B                    &    0.47$\pm$.05     &    0.50$\pm$.02     &    0.48$\pm$.02     &    0.45$\pm$.03     &    0.54$\pm$.01     &    0.46$\pm$.05     &    0.56$\pm$.02     &    0.53$\pm$.02     \\
    Biomni-R0-32B                  &    0.42$\pm$.05     &    0.55$\pm$.03     &    \underline{0.53$\pm$.01}     &    0.48$\pm$.08     &    0.55$\pm$.03     &    \textbf{0.55$\pm$.03}     &    \underline{0.58$\pm$.01}     &    0.56$\pm$.03     \\        
    \midrule
    \rowcolor{gray!10}
    \multicolumn{9}{c}{\textit{Ours}} \\
    \textbf{\methodname-8B}        &    \textbf{0.52$\pm$.02}     &    \textbf{0.69$\pm$.01}     &    \textbf{0.64$\pm$.02}     &    \textbf{0.58$\pm$.09}     &    \textbf{0.68$\pm$.03}     &    \textbf{0.55$\pm$.02}     &    0.56$\pm$.02     &    \textbf{0.81$\pm$.02}     \\
    \bottomrule
    \end{tabular*}
    \caption{\textbf{AUROC on compound perturbation QA across 8 primary organs and tissues.} Binary AUROC scores for gene regulation direction prediction across eight primary organ and tissue categories for generalist LLMs, specialist LLMs, and \benchmarkname. The best performing model for each category is marked in \textbf{bold}. Values represent the mean $\pm$ standard deviation over 3 independent runs}
    \label{tab:organ_perturbation_auroc}
\end{table*}

\paragraph{Difficulty-aware data sorting.}
Prior to inference, we optimize the learning trajectory by implementing a curriculum-based data organization strategy as described in \Cref{fig:difficulty_aware_data_ordering}. We compute a composite priority score for each sample, defined as the product of a LLM's self-consistency \citep{wang2022selfconsistency} score and biological relevance score. The biological score quantifies the interaction strength between the perturbation's MoA and the target gene using the STRING database \citep{szklarczyk2023string}, ensuring the model focuses on biologically plausible connections first. Simultaneously, the consistency score measures the stability of a standard LLM's predictions over multiple stochastic trials, effectively filtering out noisy samples where the signal is ambiguous. By sorting the dataset based on this metric, the framework prioritizes samples that are both biologically grounded and suited for reasoning, stabilizing the subsequent multi-agent optimization.

\paragraph{Progressive reasoning.}
At inference time, we further enhance reasoning by applying a progressive reasoning as described in \Cref{fig:progressive_reasoning}. Specifically, it loops over target genes conditioned on a fixed cell line $c$ and perturbation $p$. Targets are ordered by a difficulty score and processed from easier to harder, and the reasoning traces for earlier targets are stored as reusable context. For subsequent harder targets, we supply both the original biological context and a curated subset of prior reasoning, enabling evidence reuse without leaking labels. This staged conditioning reduces hallucinated causal links and improves consistency for challenging genes by anchoring them to validated reasoning from simpler cases.

\paragraph{Multi-expert reasoning.}
We deploy a collaborative multi agent architecture to synthesize a final prediction from diverse biological perspectives. First, three specialized agents, the context, mechanism, and network scientists, independently analyze the genomic background, biochemical mode of action, and signaling topology respectively. These disparate reasoning traces are then aggregated by an integration agent, which synthesizes the agents insights and knowledge to formulate a final response to the user question. Optionally, this agent can incorporate embeddings from a pretrained neural network. If integrated, LLM reasons both a standard canonical hypothesis and a counterfactual one informed by the latent signals of the network. By resolving potential conflicts between textual reasoning and neural priors, the system generates a unified candidate prediction grounded in both mechanistic logic and structured knowledge. We further details the explanation and prompt of each agents at \Cref{sec:appdx_prompts}.

\paragraph{Iterative verification and refinement.} 
We apply four judges that independently screen for history leakage, target grounding, and answer consistency. The pipeline counts problematic verdicts and retries when any judge flags an issue. Formally, we accept a reasoning trace only if the problematic count is zero.
\[
 m = \sum_{j=1}^{4} \mathbbm{1}[v_j = \text{problematic}], \quad \text{accept if } m = 0.
\]

The judges also provide natural language feedback for later analysis. This feedback helps identify systematic errors such as off-target assumptions or mismatched cell-line context. The retry loop uses the same input but seeks a more coherent reasoning chain. In practice, this filter reduces noisy decisions that stem from shallow explanations.
\section{Experiment}
\subsection{Experimental setting}

In this section, we present experimental results of three main tasks: (1) gene regulation direction prediction on \benchmarkname, (2) MoA case study on \benchmarkname\ and (3) gene regulation direction prediction on PerturbQA \citep{wu2025perturbqa}. In \benchmarkname\ we compare our performance with competitive baselines ranging from 7B to 30B parameters. Although our framework utilizes a 8B model (DeepSeek-R1-distill-Llama-8B) it demonstrates robust performance against larger-scale baselines. 

\paragraph{Baselines.} In \benchmarkname\ benchmark, we compare \methodname\ against two distinct groups of LLMs: (i) general-purpose LLMs including Llama3-8B \citep{grattafiori2024llama}, DeepSeek-R1-Distill-Llama-8B \citep{deepseekai2025deepseekr1incentivizingreasoningcapability}, Mistral-Small-3-2-24B \citep{mistral_small_2025}, and Qwen3-30B-A3B \citep{yang2025qwen3} and (ii) domain-specific LLMs including BioMistral-7B \citep{labrak2024biomistral}, BioMedGPT-LM-7B \citep{luo2023biomedgpt}, TxGemma-27B \citep{wang2025txgemma}, and Biomni-R0-32B \citep{huang2025biomni}. In PerturbQA benchmark, we compared with SUMMER \citep{wu2025perturbqa}, the SOTA model of the benchmark and followed the other baselines, \ie GAT \citep{velivckovic2017gat}, GEARS \citep{roohani2024gears}, scGPT \citep{cui2024scgpt}, GenePT \citep{chen2024genept}, as reported in the paper.

\paragraph{Metrics.} For \benchmarkname\ gene regulation direction prediction task, we report binary AUROC scores aggregated by tissue type. In mechanism of action (MoA) case study is assessed via target rank, mean gap, and relative dominance. We provide the performance of \benchmarkname\ results under alternative categories in the \Cref{sec:appdx_gene_direction}. For the PerturbQA direction of change task, we follow the original benchmark's protocol by first calculating the AUROC for target genes and subsequently reporting the mean and standard deviation across all perturbations. To ensure the reliability of our findings, we report the average performance over three independent runs. 

\subsection{\benchmarkname\ benchmark} \label{exp:lincsqa_benchmark}
In this section, we evaluate the performance of various language models on the proposed \benchmarkname\ benchmark for gene-level and MoA-level. We first present the results for gene-level regulation direction prediction, which assesses a model's ability to identify transcriptional changes under specific chemical perturbations. Subsequently, we analyze MoA prediction performance, where the models must recognize correct gene regulation direction when correct MoA and cell line pair is given. Specifically, we tested two scenarios: (1) mutation-selective inhibitors in cell lines with versus without the target mutation, and (2) mutation-selective inhibitors in cell lines harboring the target mutation but exhibiting differential drug sensitivity due to bypass resistance mechanisms.

\paragraph{Gene regulation direction prediction.}
Gene regulation direction prediction evaluates the model’s ability to classify whether a specific gene's expression is upregulated or downregulated following a compound treatment in a particular cellular context. For a given compound-cell line pair and a candidate MoA, the model performs binary classification for each target gene, predicting its regulation state as either increased (1) or decreased (0). This task tests the model's fundamental understanding of signaling cascades and downstream transcriptional consequences induced by chemical perturbations. In this setting, we incorporate the prediction and confidence of neural network, \ie Graph Attention Network \citep{velivckovic2017gat}, as a tool that \methodname\ can utilize.

The performance is reported at \Cref{tab:organ_perturbation_auroc}. To ensure biological robustness, we report the performance using binary AUROC scores aggregated over primary organ and tissue level. Notably, \benchmarkname\ consistently outperforms both general-purpose and domain-specific baselines, demonstrating that a multi-agent approach can effectively resolve entangled regulatory signals without the need for additional training. Despite its compact 8B parameter size, our framework achieves superior accuracy compared to significantly larger models, such as Qwen3-30B-A3B and Biomni-R0-32B, as well as established specialist LLMs. This performance gain suggests that progressive reasoning provides a more stable foundation for capturing complex transcriptional dependencies than monolithic architectures, even those with substantially higher parameter counts.
\begin{table*}[t]
\centering
\resizebox{0.88\textwidth}{!}{%
\begin{tabular}{l|ccc|ccc}
\hline
\multirow{2}{*}{Model} & \multicolumn{3}{c|}{Dabrafenib} & \multicolumn{3}{c}{Vemurafenib} \\
\cline{2-7}
 & \makecell{Target\\rank $\downarrow$} & \makecell{Mean\\gap $\uparrow$} & \makecell{Relative\\dominance (\%)$\uparrow$} & \makecell{Target\\rank $\downarrow$} & \makecell{Specificity\\gap $\uparrow$} & \makecell{Relative\\Dominance (\%)$\uparrow$} \\
\hline
Llama-3-8B    & 4 & -0.110 & -20.1 & 2 & 0.029 & 5.1 \\
Qwen3-30B-A3B & 5 & -0.070 & -45.7 & 2 & -0.025 & -28.1 \\
\textbf{\methodname-8B} & \textbf{1} & \textbf{0.161} & \textbf{41.3} & \textbf{1} & \textbf{0.177} & \textbf{64.6} \\
\hline
\end{tabular}%
}
\vspace{1.5mm}
\caption{\textbf{Performance on BRAF inhibitor case study.} Target rank, mean gap, and relative dominance for A375 cell line of dabrafenib and vemurafenib. We mark the best result in \textbf{bold}. We run 3 independent runs and details of metric is described in \Cref{appdx:metrics}.}
\vspace{-4mm}
\label{tab:case_study_braf}
\end{table*}
\begin{figure*}[t]
    \centering
    \includegraphics[width=\textwidth]{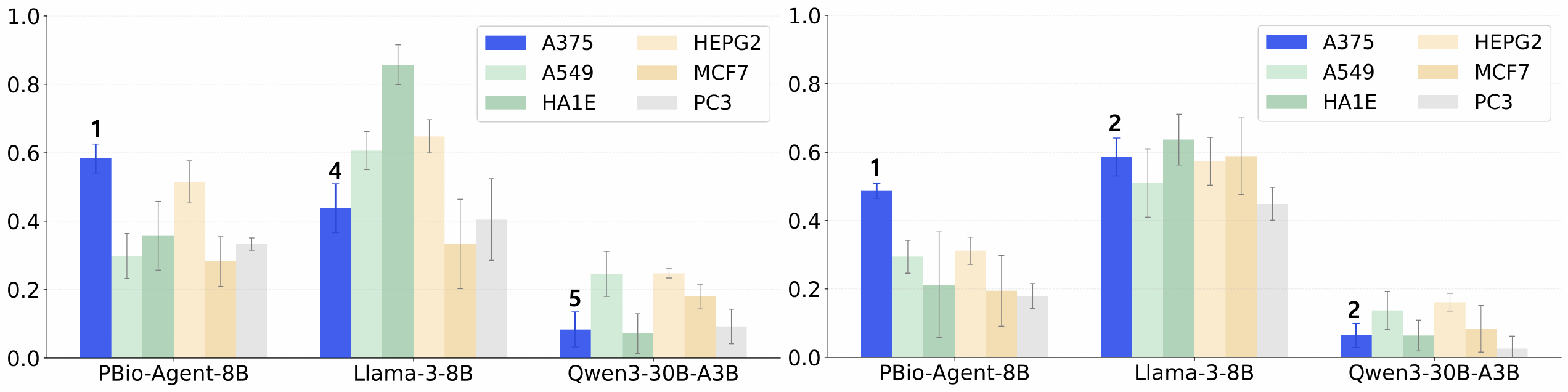}
    \vspace{-6mm}
    \caption{\textbf{Agreement ratios and target (A375 cell line) rank comparison for BRAF inhibitors.} Agreement ratios for vemurafenib (left) and dabrafenib (right) with target ranks (numbers above bars) showing A375's ranking among six cell lines. Only PBio-Agent-8B consistently achieves rank 1 in A375 (BRAF V600E-mutant), while baseline models show higher agreement in wild-type cell lines, demonstrating PBio-Agent-8B's ability to correctly prioritizes the mutation-harboring target (A375) cell line.}
    \label{fig:case_study_braf}
\end{figure*}

\begin{figure}[t]
    \centering
    \includegraphics[width=\columnwidth]{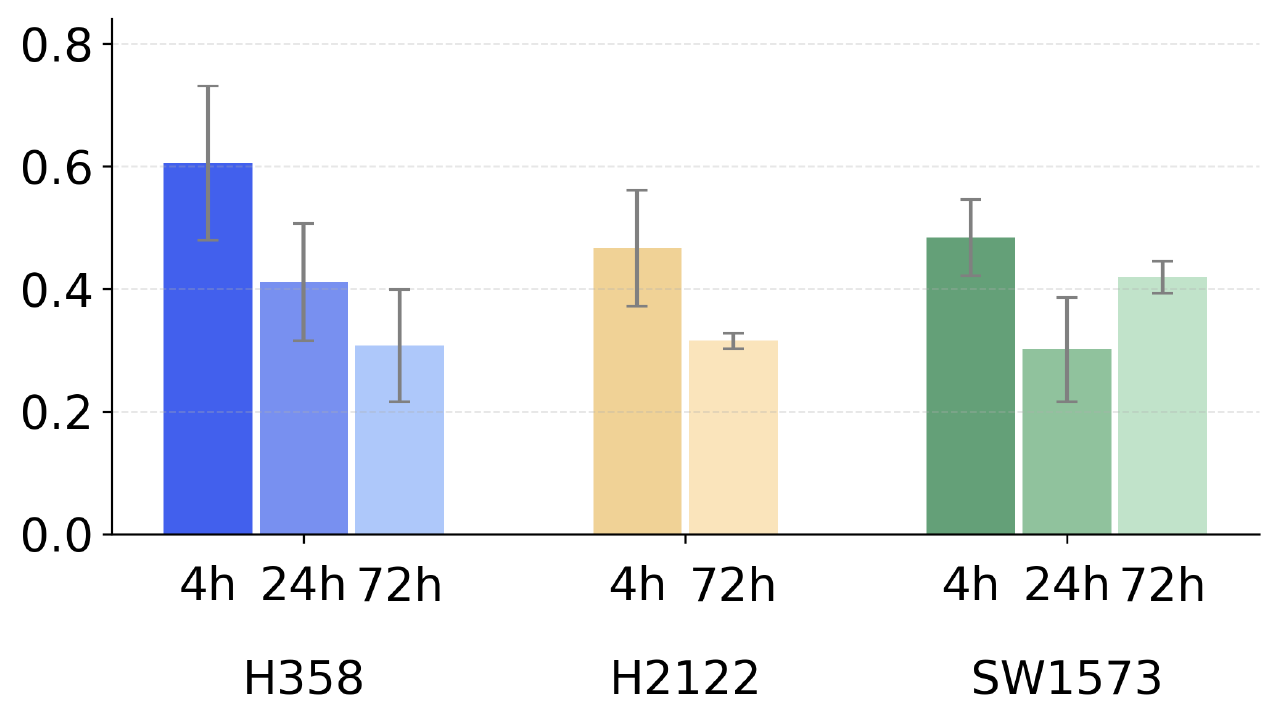}
    
    \caption{\textbf{Agreement ratios of \methodname\ across KRAS G12C-mutants with varying drug sensitivity.} H358 (sensitive), H2122 (intermediate), and SW1573 (resistant) cells were treated with ARS-1620 and evaluated at 4h, 24h, and 72h. Higher agreement in sensitive H358 reflects coherent KRAS inhibition response, while lower agreement in resistant SW1573 indicates bypass pathway activation that decouples transcriptional changes from the annotated mechanism of action.}
    \label{fig:case_study_kras}
\end{figure}

\paragraph{Case 1: BRAF V600E inhibitor (LINCS L1000).} \label{sec:case_study_1}
Gene expression signatures were extracted from the LINCS L1000 dataset \citep{duan2016lincsl1000cds2} for two BRAF V600E-selective inhibitors, vemurafenib and dabrafenib. Consensus signatures were constructed following the procedure described in \Cref{sec:method_dataset_curation}.
For the sensitive cell line, we selected A375 (melanoma), which harbors the BRAF V600E mutation with well-documented sensitivity to BRAF inhibitors.

For negative controls, we selected five BRAF wild-type cell lines from the LINCS touchstone panel: A549 (lung adenocarcinoma), MCF7 (breast cancer), PC3 (prostate cancer), HA1E (kidney), and HEPG2 (hepatocellular carcinoma). Since BRAF V600E inhibitors selectively bind to the mutant BRAF protein, these wild-type cells have no pharmacological target and thus should not produce MoA-consistent transcriptional changes.

The performance results for the BRAF inhibitor case study are reported in \Cref{tab:case_study_braf}. To evaluate cell-line specificity, we utilize three complementary metrics: target rank, mean gap (difference between A375 performance and overall mean), and relative dominance (percentage by which A375 exceeds the mean). Notably, \methodname\ achieves rank 1 for the A375 cell line in both dabrafenib and vemurafenib as described in \Cref{fig:case_study_braf}. It records mean gaps of 0.161 and 0.177 respectively, making it the only model to demonstrate positive gaps across both inhibitors. This translates to relative dominance values of 41.3\% and 64.6\%, confirming that the framework correctly prioritizes the BRAF V600E-mutant cell line over wild-type controls.

\paragraph{Case 2: KRAS G12C Inhibitor (GSE137912).} \label{sec:case_study_2}
In this task, we evaluate our \methodname\ on single-cell context. Single-cell RNA sequencing data were obtained from GSE137912, comprising KRAS G12C-mutant NSCLC cell lines treated with ARS-1620 (10 $\mu$M) at 0, 4, 24, and 72 hours. To create a robust evaluation set, we generated pseudo-bulk expression profiles by aggregating single-cell data from KRAS G12C-mutant cell lines. Ground truth labels for gene regulation were derived via differential expression analysis, identifying genes with statistically significant fold changes following drug treatment. We further describe the details of data preprocessing step at \Cref{appdx:details_kras}.

We hypothesized that despite both cell lines harboring the KRAS G12C mutation, \methodname\ should achieve higher gene-level accuracy in H358, where KRAS inhibition produces coherent downstream effects, compared to SW1573, where bypass pathway activation decouples the transcriptional response from the annotated MoA.
We visualize the agreement ratios in \Cref{fig:case_study_kras}. We evaluated \methodname\ across three KRAS G12C-mutant NSCLC cell lines with differential sensitivity to ARS-1620: H358 (sensitive), H2122 (intermediate), and SW1573 (resistant). Transcriptional responses were measured at 4, 24, and 72 hours post-treatment. At early time points (4h and 24h), agreement ratios showed a clear sensitivity-dependent gradient: H358 achieved the highest values, H2122 exhibited intermediate agreement, and SW1573 showed substantially lower concordance. This pattern directly reflects the underlying biology---H358 exhibits coherent KRAS inhibition-driven transcriptional responses, whereas SW1573's co-occurring PIK3CA mutation enables bypass pathway activation that decouples transcriptional output from the annotated mechanism of action. At 72 hours, agreement ratios declined in the sensitive and intermediate cell lines (H358 and H2122), consistent with the emergence of adaptive responses that diverge from immediate mechanistic signatures. We therefore focus our interpretation on early time points, which more faithfully capture direct MoA-driven transcriptional changes.
These results demonstrate that \methodname\ can distinguish genuine drug-target engagement from resistance-mediated pathway bypass without requiring explicit resistance annotations during inference.

\subsection{PerturbQA benchmark}
In this section, we evaluate the reasoning capabilities of \methodname\ on the PerturbQA benchmark reported at \Cref{tab:perturbqa}. Following the experimental protocols established by \citet{wu2025perturbqa}, we conduct a comparative analysis against various baseline models to assess their proficiency in predicting transcriptional responses to genetic perturbations. Gene regulation direction prediction within the PerturbQA framework requires models to identify whether a target gene is upregulated or downregulated following a specific genetic perturbation.

As shown in the \Cref{tab:perturbqa}, \methodname\ demonstrates robust performance across multiple cell lines, including K562, RPE1, HepG2, and Jurkat. Specifically, our framework achieves state-of-the-art results on HepG2 (0.67) and Jurkat (0.68). While models like GEARS and SUMMER show competitive performance in specific contexts such as K562 and RPE1, \methodname\ maintains superior or comparable accuracy across the entire suite of tested cell lines. These results indicate that integrating multi-agent reasoning with structured biological knowledge effectively captures the complex dependencies inherent in genetic perturbation data.

\begin{table}[t]
\centering
\small
\begin{tabular}{lllll}
\toprule
Model & K562 & RPE1 & HepG2 & Jurkat \\ \midrule
\textsc{Gat} & 0.58 & \underline{0.60} & \underline{0.64} & 0.59 \\
\textsc{Gears} & \textbf{0.64} & \underline{0.60} & 0.52 & 0.51 \\
\textsc{ScGpt} & 0.48 & 0.53 & 0.51 & 0.51 \\ 
\textsc{GenePt-Gene} & 0.53 & 0.57 & 0.58 & 0.57 \\
\textsc{GenePt-Prot} & 0.57 & 0.57 & 0.55 & 0.58 \\
LLM (No CoT) & 0.50 & 0.49 & 0.49 & 0.50 \\
LLM (No retrieval) & 0.49 & 0.52 & 0.51 & 0.53 \\
Retrieval (No LLM) & 0.50 & 0.50 & 0.50 & 0.50 \\
\textsc{Summer} & \underline{0.62} & \textbf{0.64} & \underline{0.65} & \underline{0.66} \\ \cmidrule{1-5} 
\textbf{\methodname-8B} & \textbf{0.64} & \underline{0.60} & \textbf{0.67} & \textbf{0.68} \\ \bottomrule
\end{tabular}
\vspace{2mm}
\caption{\textbf{Performance of direction of change of PerturbQA.} We report AUROC over 3 runs by following original protocol of PerturbQA.}
\label{tab:perturbqa}
\end{table}
\section{Related Works}
\paragraph{Benchmarks for cellular perturbation.}
The emergence of giga-scale atlases has provided a robust foundation for evaluating model fidelity across diverse chemical and genetic landscapes. Tahoe-100M \citep{zhang2025tahoe100m} study compiled a massive perturbation atlas covering 92 cancer cell lines and thousands of compounds, enabling deep analysis of context-dependent gene functions. To address data consistency issues across disparate studies, scBaseCount \citep{youngblut2025scbasecount} utilized AI agents to uniformly reprocess vast single-cell datasets, establishing a unified pipeline that minimizes analytical variability from different reference genomes. 

Perturb-seq \citep{dixit2016perturbseq} revolutionized the single-cell gene perturbation studies by combining pooled genetic screening with single-cell RNA sequencing. Recently, \citet{wu2025perturbqa} introduced a single-cell gene perturbation benchmark for language models. In this work, we present a benchmark for bulk-cell compound perturbations. While critical for drug discovery, this area remains under-explored, and the reasoning capabilities of LLMs in this context have not yet been fully investigated.
\vspace{-2mm}
\paragraph{Language models for cellular perturbation.}
Early studies applied the language models by pre-training them on large-scale transcriptomic data. The studies view gene expression profiles as \textit{sentences} and learn the hidden structure of gene networks \citep{yang2022scbert,cui2024geneformer,cui2024scgpt}. Alongside these data-driven models, LLMs focused on biomedical literature which can synthesize a lot of prior biological knowledge \citep{luo2022biogpt, luo2023biomedgpt}. They predict and explain complex molecular interactions effectively. 

Building on this groundwork, PerturbQA \citep{wu2025perturbqa} benchmarks structured reasoning over experimental single-cell genetic perturbation data. PerturbQA demonstrates LLMs' potential for interpretable reasoning by focusing on causal relationships rather than retrieval. However, unlike PerturbQA's independent target gene prediction, our framework jointly analyzes target gene sets affected by the same perturbation. Furthermore, our confidence-aware ordering propagates high-confidence information to inform subsequent steps.

\vspace{-1mm}
\paragraph{Curriculum-driven inference.}
Curriculum-based data ordering enables prior knowledge to guide subsequent predictions. This concept originates from curriculum learning (CL), proposed by \citep{bengio2009curriculum}. This structured data ordering concept is now actively employed in LLMs. \citet{kim2024strategicdataordering} demonstrated that metric-based ordering outperforms simple length-based sorting. Moreover, curriculum-based sequencing has been shown to facilitate domain mastery \citep{neema2025acerfromamateur, yang2024medicalqaevaluating} and optimize diverse capabilities, including preference alignment, multilingual proficiency, and convergence speed \citep{zhang2025preferencecurriculum, pucci2023arealllanguagesequal, zhang2506beyondrandom}. 

Leveraging these insights, our work adopts this progressive strategy for cell perturbation analysis by sequentially accumulating predicted results at inference time. This mirrors biological reality, where analyzing gene regulations step-by-step builds a more accurate understanding of cellular responses. Unlike training-time curriculum learning, our approach implements difficulty-aware ordering at inference time, dynamically prioritizing genes based on prediction confidence and biological relatedness without requiring model retraining.

\section{Conclusion}

We present \methodname\ for bulk-cell compound perturbation. The method integrates multi-sci reasoning, knowledge graph context, and judge-based verification in a single pipeline. This design enforces structured correlations and reduces brittle inference that is common in single-pass LLM systems. Our evaluation demonstrates the feasibility of combining agentic reasoning with biological constraints for both direction prediction and MoA selection. Larger multi-omics datasets and richer pathway resources will be needed to scale the method. Future work should explore lighter judge ensembles and stronger biological retrieval.

\section*{Impact Statement} 
This work introduces \benchmarkname, a novel benchmark for evaluating the reasoning capabilities of large language models (LLMs) in predicting transcriptional responses to chemical perturbations in bulk-cell environments. By proposing \methodname, a multi-agent framework that utilizes progressive reasoning and structured biological knowledge, this research enhances the accuracy and interpretability of computational drug discovery. The methodology does not involve human participants or the use of sensitive personal data, and thus we do not anticipate direct ethical risks from the experiments presented. However, the development of more reliable AI systems for predicting biological causalities may significantly accelerate the discovery of new therapeutics and the understanding of disease mechanisms. We advocate for the responsible application of these methods in drug development and encourage practitioners to adhere to established safety, legal, and professional guidelines to mitigate potential societal risks associated with automated biological reasoning.
\bibliography{example_paper}
\bibliographystyle{icml2026}

%%%%%%%%%%%%%%%%%%%%%%%%%%%%%%%%%%%%%%%%%%%%%%%%%%%%%%%%%%%%%%%%%%%%%%%%%%%%%%%
%%%%%%%%%%%%%%%%%%%%%%%%%%%%%%%%%%%%%%%%%%%%%%%%%%%%%%%%%%%%%%%%%%%%%%%%%%%%%%%
% APPENDIX
%%%%%%%%%%%%%%%%%%%%%%%%%%%%%%%%%%%%%%%%%%%%%%%%%%%%%%%%%%%%%%%%%%%%%%%%%%%%%%%
%%%%%%%%%%%%%%%%%%%%%%%%%%%%%%%%%%%%%%%%%%%%%%%%%%%%%%%%%%%%%%%%%%%%%%%%%%%%%%%
\newpage
\appendix
\onecolumn

\section{Dataset}\label{appdx:dataset}
\begin{table}[h]
\centering

\label{tab:dataset_statistics}
\begin{tabular}{llrrrr}
\midrule
\multirow{2}{*}{Dataset} & \multirow{2}{*}{Split} & \multirow{2}{*}{Total} & \multicolumn{3}{c}{Differentially expressed} \\
\cline{4-6}
 & & & Total & Up & Down \\
 \midrule
 \midrule

Bone marrow & Train & 259 & 259 & 125 & 134 \\
 & Test & 97 & 97 & 59 & 38 \\
 \midrule
Breast & Train & 1,110 & 1,110 & 617 & 493 \\
 & Test & 390 & 390 & 175 & 215 \\
 \midrule
Cervix & Train & 694 & 694 & 404 & 290 \\
 & Test & 296 & 296 & 131 & 165 \\
 \midrule
Colon & Train & 267 & 267 & 126 & 141 \\
 & Test & 80 & 80 & 47 & 33 \\
 \midrule
Lung & Train & 449 & 449 & 248 & 201 \\
 & Test & 156 & 156 & 79 & 77 \\
 \midrule
Peripheral blood & Train & 498 & 498 & 253 & 245 \\
 & Test & 182 & 182 & 97 & 85 \\
 \midrule
Prostate & Train & 270 & 270 & 127 & 143 \\
 & Test & 124 & 124 & 71 & 53 \\
 \midrule
Skin & Train & 653 & 653 & 327 & 326 \\
 & Test & 227 & 227 & 117 & 110 \\
\bottomrule
\end{tabular}
\caption{Statistics of the gene regulation direction  benchmark of \benchmarkname\ across organ-specific category.}
\end{table}

\subsection{Details of KRAS G12C inhibitor data preprocessing}
\label{appdx:details_kras}
Raw count matrices were processed to generate pseudobulk expression profiles by aggregating single-cell data within each cell line and time point. Differential expression analysis was performed comparing treated samples (4, 24 and 72 hours) to untreated controls (0 hours) using the Mann-Whitney U test. Differentially expressed genes were identified using the following criteria: false discovery rate (FDR) $<$ 0.05 and $|\log_2 \text{ fold change}| > 0.5$. The direction of regulation was determined by the sign of the log2 fold change for each gene passing these thresholds.
All three cell lines in GSE137912 harbor KRAS G12C mutations but exhibit differential sensitivity to G12C-selective inhibitors. H358 is classified as sensitive (sotorasib IC50 ~0.13 µM), H2122 exhibits intermediate sensitivity, and SW1573 is classified as resistant (IC50 ~9.6 µM) due to a co-occurring PIK3CA mutation that enables bypass pathway activation.
We hypothesized that despite both cell lines harboring the KRAS G12C mutation, \methodname\ should achieve higher gene-level accuracy in H358, where KRAS inhibition produces coherent downstream effects, compared to SW1573, where bypass pathway activation decouples the transcriptional response from the annotated MoA.
\section{Metrics}\label{appdx:metrics}
\subsection{Case Study Evaluation Metrics} 
In our BRAF V600E inhibitor case study (\Cref{sec:case_study_1}), we evaluate cell-line specificity using three complementary metrics that quantify how well a model prioritizes the target cell line (A375, BRAF V600E-mutant) over wild-type cell lines.

\paragraph{Notation.}
Let $\mathcal{C} = \{c_1, c_2, \ldots, c_K\}$ denote the set of $K$ cell lines, where $c_{\text{target}} = \text{A375}$ is the target cell line harboring the BRAF V600E mutation. For a given drug $d$ (dabrafenib or vemurafenib) and cell line $c_i$, let $r_{d,c_i}$ denote the gene-level agreement ratio—the proportion of genes where the model correctly predicts the direction of regulation.

\paragraph{Target Rank}
The target rank measures the ranking position of the target cell line among all $K$ cell lines, based on agreement ratios.

\begin{equation}
\text{Target Rank} = 1 + \sum_{c_i \in \mathcal{C} \setminus \{c_{\text{target}}\}} \mathbbm{1}[r_{d,c_i} > r_{d,c_{\text{target}}}]
\end{equation}

where $\mathbbm{1}[\cdot]$ is the indicator function. A target rank of 1 indicates that the target cell line has the highest agreement ratio among all cell lines, reflecting optimal cell-line specificity. Higher ranks indicate that the model achieves better performance on wild-type cell lines than on the mutation-harboring target, suggesting a failure to capture target-mutation-dependent transcriptional responses.

\paragraph{Mean Gap (Specificity Gap)}
The mean gap, also referred to as specificity gap, quantifies the absolute difference between the target cell line's performance and the average performance across all cell lines.

\begin{equation}
\text{Mean Gap} = r_{d,c_{\text{target}}} - \frac{1}{K} \sum_{c_i \in \mathcal{C}} r_{d,c_i}
\end{equation}

A positive mean gap indicates that the target cell line outperforms the average, reflecting genuine target specificity. A negative gap suggests that the model achieves higher agreement in off-target (wild-type) cell lines, indicating a lack of mutation-aware prediction capability. This metric is particularly informative because it accounts for overall model performance: a model might achieve high absolute agreement ratios across all cell lines but still fail to prioritize the pharmacologically relevant target.

\paragraph{Relative Dominance}
Relative dominance expresses the mean gap as a percentage of the overall mean performance, providing a scale-normalized measure of target specificity.

\begin{equation}
\text{Relative Dominance (\%)} = \frac{r_{d,c_{\text{target}}} - \bar{r}_d}{\bar{r}_d} \times 100
\end{equation}

where $\bar{r}_d = \frac{1}{K} \sum_{c_i \in \mathcal{C}} r_{d,c_i}$ is the mean agreement ratio across all cell lines for drug $d$. This metric allows for comparison across models with different baseline performance levels. A positive relative dominance indicates that the target cell line's performance exceeds the mean by the specified percentage, while negative values indicate underperformance relative to the average.

\paragraph{Interpretation}
These three metrics collectively provide a comprehensive assessment of cell-line specificity:
\begin{itemize}
    \item \textbf{Target Rank} directly measures whether the target cell line is prioritized (rank 1 is optimal).
    \item \textbf{Mean Gap} quantifies the magnitude of target cell line advantage in absolute terms.
    \item \textbf{Relative Dominance} normalizes this advantage relative to overall performance, enabling fair comparison across models with different baseline capabilities.
\end{itemize}

In \Cref{tab:case_study_braf}, we observe that \methodname\ is the only model achieving rank 1 with positive mean gaps and substantial relative dominance values across both BRAF inhibitors, demonstrating robust target-mutation-aware prediction capability.
\clearpage
\newpage
\section{Gene regulation direction prediction}
\label{sec:appdx_gene_direction}
\subsection{Cancer type}

\begin{table}[H]
\centering
\footnotesize
\renewcommand{\arraystretch}{1.2}
\newcolumntype{Y}{>{\raggedright\sloppy\arraybackslash}X}
\begin{tabularx}{\textwidth}{@{} Y *{8}{c} @{}}
\toprule
\textbf{Model} & \textbf{Breast} & \textbf{Cervical} & \textbf{Leukemia} & \textbf{Melanoma} & \textbf{Lung} & \textbf{Prostate} & \textbf{Colorectal} & \textbf{Lymphoma} \\
\midrule
\rowcolor{gray!10}
\multicolumn{9}{c}{\textit{Generalist LLM}} \\
Llama3-8B                       &    0.53$\pm$0.00    &    0.49$\pm$0.00    &    0.50$\pm$0.06    &    0.53$\pm$0.00    &    \underline{0.58$\pm$0.00}    &    0.50$\pm$0.00    &    \underline{0.53$\pm$0.00}    &    0.50$\pm$0.06    \\
DeepSeek-R1-Distill-Llama-8B   &    0.49$\pm$0.00    &    0.51$\pm$0.00    &    \underline{0.51$\pm$0.04}    &    0.50$\pm$0.00    &    0.50$\pm$0.00    &    0.52$\pm$0.00    &    0.46$\pm$0.00    &    \underline{0.51$\pm$0.04}    \\
Mistral-Small-3.2-24B          &    0.52$\pm$0.00    &    0.50$\pm$0.00    &    0.46$\pm$0.07    &    0.53$\pm$0.00    &    0.55$\pm$0.00    &    0.52$\pm$0.00    &    0.47$\pm$0.00    &    0.46$\pm$0.07    \\
Qwen3-30B-A3B                  &    \underline{0.62$\pm$0.00}    &    \underline{0.53$\pm$0.00}    &    0.50$\pm$0.04    &    \underline{0.60$\pm$0.00}    &    \underline{0.58$\pm$0.00}    &    \textbf{0.66$\pm$0.00}    &    0.43$\pm$0.00    &    0.50$\pm$0.04    \\
\midrule
\rowcolor{gray!10}
\multicolumn{9}{c}{\textit{Specialist LLM}} \\
BioMistral-7B                  &    0.51$\pm$0.00    &    0.49$\pm$0.00    &    0.48$\pm$0.02    &    0.51$\pm$0.00    &    0.49$\pm$0.00    &    0.50$\pm$0.00    &    0.51$\pm$0.00    &    0.48$\pm$0.02    \\
BioMedGPT-LM-7B                &    0.50$\pm$0.00    &    0.50$\pm$0.00    &    0.50$\pm$0.00    &    0.50$\pm$0.00    &    0.50$\pm$0.00    &    0.50$\pm$0.00    &    0.50$\pm$0.00    &    0.50$\pm$0.00    \\
TxGemma-27B                    &    0.50$\pm$0.00    &    0.48$\pm$0.00    &    0.47$\pm$0.01    &    0.53$\pm$0.00    &    0.54$\pm$0.00    &    0.56$\pm$0.00    &    0.45$\pm$0.00    &    0.47$\pm$0.01    \\
Biomni-R0-32B                  &    0.55$\pm$0.00    &    \underline{0.53$\pm$0.00}    &    0.48$\pm$0.09    &    0.56$\pm$0.00    &    0.55$\pm$0.00    &    \underline{0.58$\pm$0.00}    &    0.48$\pm$0.00    &    0.48$\pm$0.09    \\
\midrule
\rowcolor{gray!10}
\multicolumn{9}{c}{\textit{Ours}} \\
\textbf{\methodname-8B}               &    \textbf{0.69$\pm$0.00}    &    \textbf{0.64$\pm$0.00}    &    \textbf{0.53$\pm$0.02}    &    \textbf{0.81$\pm$0.00}    &    \textbf{0.68$\pm$0.00}    &    0.56$\pm$0.00    &    \textbf{0.58$\pm$0.00}    &    \textbf{0.53$\pm$0.02}    \\
\bottomrule
\end{tabularx}
\caption{AUROC on compound perturbation QA across 8 cancer types.}
\label{appdx:tab:cancer_type_auroc}
\end{table}
\subsection{Solid vs Hematological}
\begin{table}[H]
\centering
\small
\renewcommand{\arraystretch}{1.2}
\begin{tabularx}{0.5\textwidth}{@{} X *{2}{c} @{}}
\toprule
\textbf{Model} & \textbf{Solid} & \textbf{Hematological} \\

\midrule
\rowcolor{gray!10}
\multicolumn{3}{c}{\textit{Generalist LLM}} \\
Llama3-8B                       &   0.53$\pm$0.03    &   0.50$\pm$0.06    \\
DeepSeek-R1-Distill-Llama-8B   &   0.50$\pm$0.02    &   \underline{0.51$\pm$0.04}    \\
Mistral-Small-3.2-24B          &   0.51$\pm$0.03    &   0.46$\pm$0.07    \\
Qwen3-30B-A3B                  &   \underline{0.57$\pm$0.08}    &   0.50$\pm$0.04    \\
\midrule
\rowcolor{gray!10}
\multicolumn{3}{c}{\textit{Specialist LLM}} \\
BioMistral-7B                  &   0.50$\pm$0.01    &   0.48$\pm$0.02    \\
BioMedGPT-LM-7B                &   0.50$\pm$0.00    &   0.50$\pm$0.00    \\
TxGemma-27B                    &   0.51$\pm$0.04    &   0.47$\pm$0.01    \\
Biomni-R0-32B                  &   0.54$\pm$0.04    &   0.48$\pm$0.09    \\
\midrule
\rowcolor{gray!10}
\multicolumn{3}{c}{\textit{Ours}} \\
\textbf{\methodname-8B}      &   \textbf{0.66$\pm$0.09}    &   \textbf{0.53$\pm$0.02}    \\
\bottomrule
\end{tabularx}
\caption{AUROC on compound perturbation QA across solid and hematological entities.}
\label{appdx:tab:sold_hematological}
\end{table}

\clearpage
\newpage
\section{Agents prompts}
\vspace{-4mm}
\label{sec:appdx_prompts}
\begin{table}[H]
\caption{\textbf{Prompts of agents in \methodname}} \label{tab:appdx_prompt_1}
\centering
\scalebox{0.88}{
\begin{tabular}{l m{0.77\textwidth}}
\toprule
Context agent & You are a Cancer Dependency expert. Analyze the genomic landscape of the cell line. Your role is to provide the biological 'ground' for the perturbation. \\ \addlinespace
& \textbf{OUTPUT FORMAT (STRICT - JSON ONLY):} \\
& \{ \\
& \quad "context\_reasoning": "...", \\
& \quad "pathway\_activity": "active/inactive/unknown" \\
& \} \\ \addlinespace
& \textbf{RULES:} \\
& 1) Focus on: Basal expression of target/perturb genes and key driver mutations (e.g., BRAF V600E). \\
& 2) If the target gene is not expressed, it cannot be downregulated further. \\
& 3) Use ONLY biological facts related to the specific cell line. \\ \addlinespace
& \textbf{USER PROMPT:} \\
& Analyze context: Cell Line: \{cell\_line\}, Perturbation: \{pert\_or\_moa\}, Target Gene: \{target\_gene\} \\

\midrule
Network agent & You are a Systems Biology expert. Trace the regulatory path from the perturbation target to the gene of interest. \\ \addlinespace
& \textbf{OUTPUT FORMAT (STRICT - JSON ONLY):} \\
& \{ \\
& \quad "network\_reasoning": "Step-by-step pathway reasoning using (Gene)-(rel)->(Entity) format", \\
& \quad "edge\_type": "positive\_regulation/negative\_regulation/complex" \\
& \} \\ \addlinespace
& \textbf{RULES:} \\
& 1) Trace paths: (PerturbationTarget) -(relationship)-> (Intermediate) -(relationship)-> (TargetGene) \\
& 2) Distinguish between 'Activity change' and 'Expression change'. \\
& 3) Identify feedback loops or compensatory mechanisms. \\
& 4) Use biological knowledge graph's pathway context if provided. \\ \addlinespace
& \textbf{USER PROMPT:} \\
& Trace the network path: \\
& - Start Point (Perturbation Target): \{pert\_target\} \\
& - End Point (Target Gene): \{target\_gene\} \\
& Is there a known transcriptional or signaling link between these nodes? \\
\midrule
Mechanism agent & You are a Molecular Pharmacologist. Define the immediate molecular consequence of the perturbation. \\ \addlinespace
& \textbf{OUTPUT FORMAT (STRICT - JSON ONLY):} \\
& \{ \\
& \quad "mechanism\_reasoning": "Direct effect using (Gene)-(rel)->(Entity)", \\
& \quad "primary\_action": "repression/inhibition/activation/etc" \\
& \} \\ \addlinespace
& \textbf{USER PROMPT:} \\
& Define the mechanism of action: \\
& - Perturbation: \{pert\_or\_moa\} \\
& - Chemical Name (Optional): \{drug\_name\} \\
& - Target Gene: \{target\_gene\} \\
& What is the first biochemical event that happens upon this perturbation? \\

\bottomrule
\end{tabular}}
\end{table}

\begin{table}[t]
\caption{\textbf{Prompts of agents in \methodname}} \label{tab:appdx_prompt_2}
\centering
\scalebox{0.88}{
\begin{tabular}{l m{0.77\textwidth}}
\toprule
Integration agent & You are a Molecular Biology Expert. Integrate evidence from Context, Mechanism, and Network agents to predict the target gene mRNA change. \\ \addlinespace
& \textbf{OUTPUT FORMAT (STRICT - JSON ONLY):} \\
& \{ \\
& \quad "reasoning": "Integrated pathway-grounded reasoning", \\
& \quad "answer": "upregulated/downregulated" \\
& \} \\ \addlinespace
& \textbf{DECISION STEPS:} \\
& Step 0: Summarize Agent Evidence with pathway notation. \\
& Step 1: Check for direct transcriptional evidence. \\
& Step 2: Justify UP vs DOWN case using (Gene)-(rel)->(Entity). \\
& Step 3: Final decision based on the most anchored path. \\ \addlinespace
& \textbf{USER PROMPT:} \\
& In \{cell\_line\}, will \{target\_gene\} be upregulated or downregulated by \{pert\_or\_moa\}? \\
& [Agent Evidence] Context: \{context\_reasoning\}, Mechanism: \{mechanism\_reasoning\}, Network: \{network\_reasoning\} \\
\midrule
History leakage checker & You are a History Leakage Inspector. \\
& Your ONLY task is to detect whether the reasoning relies on previous history direction labels WITHOUT introducing a new, case-specific justification. \\
& Check ONLY the following: \\
& 1) Does the reasoning explicitly or implicitly copy the direction (up/down) from prior cases? \\
& 2) Is the final direction justified by perturbation-specific reasoning, or merely by similarity to previous genes? \\ \addlinespace
& \textbf{OUTPUT FORMAT (STRICT - JSON ONLY):} \\
& \{ \\
& \quad "verdict": "problematic" or "not-problematic", \\
& \quad "feedback": "..." \\
& \} \\ \addlinespace
& \textbf{RULES:} \\
& - Using history as contextual background is ALLOWED. \\
& - Using history direction as the primary or sole justification is NOT allowed. \\
& - If history leakage is detected, verdict MUST be "problematic". \\ \addlinespace
& \textbf{USER PROMPT:} \\
& Previous History Summary: \{history\_summary\} \\
& Canonical Reasoning: \{canonical\_reasoning\} \\
& Counterfactual Reasoning: \{counterfactual\_reasoning\} \\
& Final Reasoning: \{final\_reasoning\} \\
& Final Answer: \{final\_answer\} \\

\bottomrule
\end{tabular}}
\end{table}

\clearpage
\newpage
\begin{table}[h]
\caption{\textbf{Prompts of agents in \methodname}} \label{tab:appdx_prompt_3}
\centering
\scalebox{0.88}{
\begin{tabular}{l m{0.77\textwidth}}
\toprule
Target grounding checker & You are a Grounding Consistency Inspector. \\
& Your ONLY task is to verify whether the reasoning is properly grounded in the provided biological entities. \\ \addlinespace
& \textbf{Check ONLY:} \\
& 1) Consistent reference to the given cell line? \\
& 2) Correct reference to the perturbation (gene or MoA)? \\
& 3) Correct and consistent reference to the target gene? \\
& 4) Avoidance of unrelated cell lines, genes, or drugs? \\ \addlinespace
& \textbf{OUTPUT FORMAT (STRICT - JSON ONLY):} \\
& \{ \\
& \quad "verdict": "problematic" or "not-problematic", \\
& \quad "feedback": "..." \\
& \} \\ \addlinespace
& \textbf{RULES:} \\
& - Penalize ONLY explicit mismatches or hallucinated entities. \\
& - Do NOT judge biological correctness or the final answer. \\ \addlinespace
& \textbf{USER PROMPT:} \\
& Inputs: Cell Line: \{cell\_line\}, Perturbation: \{pert\_or\_moa\}, Target Gene: \{target\_gene\} \\
& Canonical Reasoning: \{canonical\_reasoning\} \\
& Counterfactual Reasoning: \{counterfactual\_reasoning\} \\
& Final Reasoning: \{final\_reasoning\} \\
& Final Answer: \{final\_answer\} \\
\midrule
Answer consistency checker & You are a Logical Consistency Checker. \\
& Your ONLY task is to verify consistency between the reasoning text and the final answer. \\ \addlinespace
& \textbf{Check ONLY:} \\
& 1) Does the reasoning argue for upregulation while the answer says downregulated? \\
& 2) Does the reasoning argue for downregulation while the answer says upregulated? \\
& 3) Is the final answer unsupported or contradicted by the reasoning? \\ \addlinespace
& \textbf{OUTPUT FORMAT (STRICT - JSON ONLY):} \\
& \{ \\
& \quad "verdict": "problematic" or "not-problematic", \\
& \quad "feedback": "..." \\
& \} \\ \addlinespace
& \textbf{RULES:} \\
& - Do NOT judge biological validity / grounding / history usage. \\
& - If ANY inconsistency is found, verdict MUST be "problematic". \\ \addlinespace
& \textbf{USER PROMPT:} \\
& Canonical Reasoning: \{canonical\_reasoning\} \\
& Counterfactual Reasoning: \{counterfactual\_reasoning\} \\
& Final Reasoning: \{final\_reasoning\} \\
& Final Answer: \{final\_answer\} \\
& Is the reasoning logically consistent with the answer? \\
\bottomrule
\end{tabular}}
\end{table}

%%%%%%%%%%%%%%%%%%%%%%%%%%%%%%%%%%%%%%%%%%%%%%%%%%%%%%%%%%%%%%%%%%%%%%%%%%%%%%%
%%%%%%%%%%%%%%%%%%%%%%%%%%%%%%%%%%%%%%%%%%%%%%%%%%%%%%%%%%%%%%%%%%%%%%%%%%%%%%%

\end{document}